\documentclass[runningheads]{llncs}

\usepackage{eccv}

\usepackage{eccvabbrv}

\usepackage{graphicx}
\usepackage{booktabs}
\usepackage{subcaption}
\usepackage{wrapfig}
\usepackage[accsupp]{axessibility}  

\usepackage{multirow}
\usepackage{amsmath}
\usepackage{amsfonts}
\usepackage{tikz}
\usepackage{pgfplots}
\pgfplotsset{compat=1.18}
\usepackage{amssymb}
\usepackage{xcolor,colortbl}
\definecolor{lightgray}{gray}{0.9}
\usepackage{makecell}
\usepackage{diagbox}
\usepackage{subcaption}
\usepackage{algorithm}
\usepackage{algpseudocode}
\usepackage{verbatim}

\usepackage{hyperref}

\usepackage{orcidlink}

\newcommand\samethanks[1][\value{footnote}]{\footnotemark[#1]}

\begin{document}

\title{MixTTA: Low-Rank Cross-Channel Mixing for Reliable Test-Time Adaptation} 

\titlerunning{MixTTA: Low-Rank Cross-Channel Mixing for Reliable TTA}

\author{Mansoo Jung\inst{1}\orcidlink{0009-0003-7846-6521} \and
Youngwook Kim\inst{2}\thanks{Corresponding authors.}\orcidlink{0009-0005-5722-7286} \and
Jungwoo Lee\inst{1, 3}\samethanks\orcidlink{0000-0002-6804-980X}}

\authorrunning{M. Jung et al.}

\institute{
Seoul National University, Seoul, Republic of Korea \\ 
\email{\{tlwh1179, junglee\}@snu.ac.kr}
\and
Kookmin University, Seoul, Republic of Korea \\
\email{youngwook@kookmin.ac.kr}
\and
HodooAI Labs, Seoul, Republic of Korea \\
}

\maketitle
\begin{abstract}
  Test-Time Adaptation (TTA) methods commonly update the affine parameters of normalization layers to adapt deployed models under distribution shifts. However, per-channel affine parameters perform axis-aligned scaling and shifting, making them geometrically incapable of correcting cross-channel structural changes induced by distribution shift. To address this limitation, we propose MixTTA, a lightweight plug-in module that equips normalization layers with a low-rank cross-channel transformation, enabling inter-channel mixing at each layer. To ensure that the low-rank branch captures only cross-channel interactions, we also propose Decoupling Projection that enforces strict separation from the diagonal affine path, along with Spectral Projection that prevents rank-1 collapse under non-stationary test streams. MixTTA can be seamlessly integrated into any existing normalization-based TTA method. Experiments in both standard and wild TTA settings show consistent improvements over strong baselines while mitigating adaptation failure under challenging conditions. The source code is publicly available at \url{https://github.com/delta6189/MixTTA}.

\end{abstract}

\section{Introduction}
\label{sec:intro}

Deep neural networks have achieved remarkable progress over the past decade, demonstrating exceptional performance across a wide range of tasks in computer vision \cite{dosovitskiy2020image, kim2024instance, kim2025leveraging, shin2024learning, shin2025plug, kwon2024rl, he2017mask, um2025boost, lee2025dimension}.
However, most models are trained under the unrealistic assumption that the training and test distributions are identical. 
In real-world scenarios, distribution shifts, arising from environmental changes, sensor noise, or data corruption, can lead to severe performance degradation~\cite{hendrycks2019benchmarking}. 
Consequently, addressing domain shifts has become a central challenge in modern machine learning, motivating extensive research on adaptation frameworks~\cite{ganin2015unsupervised, sun2016deepcoral, saito2018maximum, cha2021swad, shi2022gradient, liang2020we, yang2021exploiting, wang2020tent}. 
Among these, test-time adaptation (TTA)~\cite{wang2020tent, iwasawa2021t3a, jung2025edas, zhang2022memo, wang2022continual, lee2024becotta, park2026rethinking, fan2026moetta} has emerged as a promising direction since it allows a deployed model to adjust itself online without access to source data or labels.


One of the foundational methods of TTA is Tent\cite{wang2020tent}, which sharpens predictions by minimizing prediction entropy while updating only the affine parameters of normalization layers. Its popularity stems from the fact that constraining updates to per-channel scale and bias maintains a compact update space, stabilizing adaptation and reducing the risk of overfitting. 
Subsequent works have typically retained the same set of learnable parameters, incorporating additional strategies such as regularization \cite{niu2022eata}, sample selection \cite{niu2023sar, lee2024entropy}, or region-level confidence estimation \cite{hu2025beyond}. Despite these advances, the structural form of the affine transformation itself has remained largely unexamined.


\begin{figure*}[t]
  \centering
  \begin{subfigure}[b]{0.42\linewidth}
      \includegraphics[width=\linewidth]{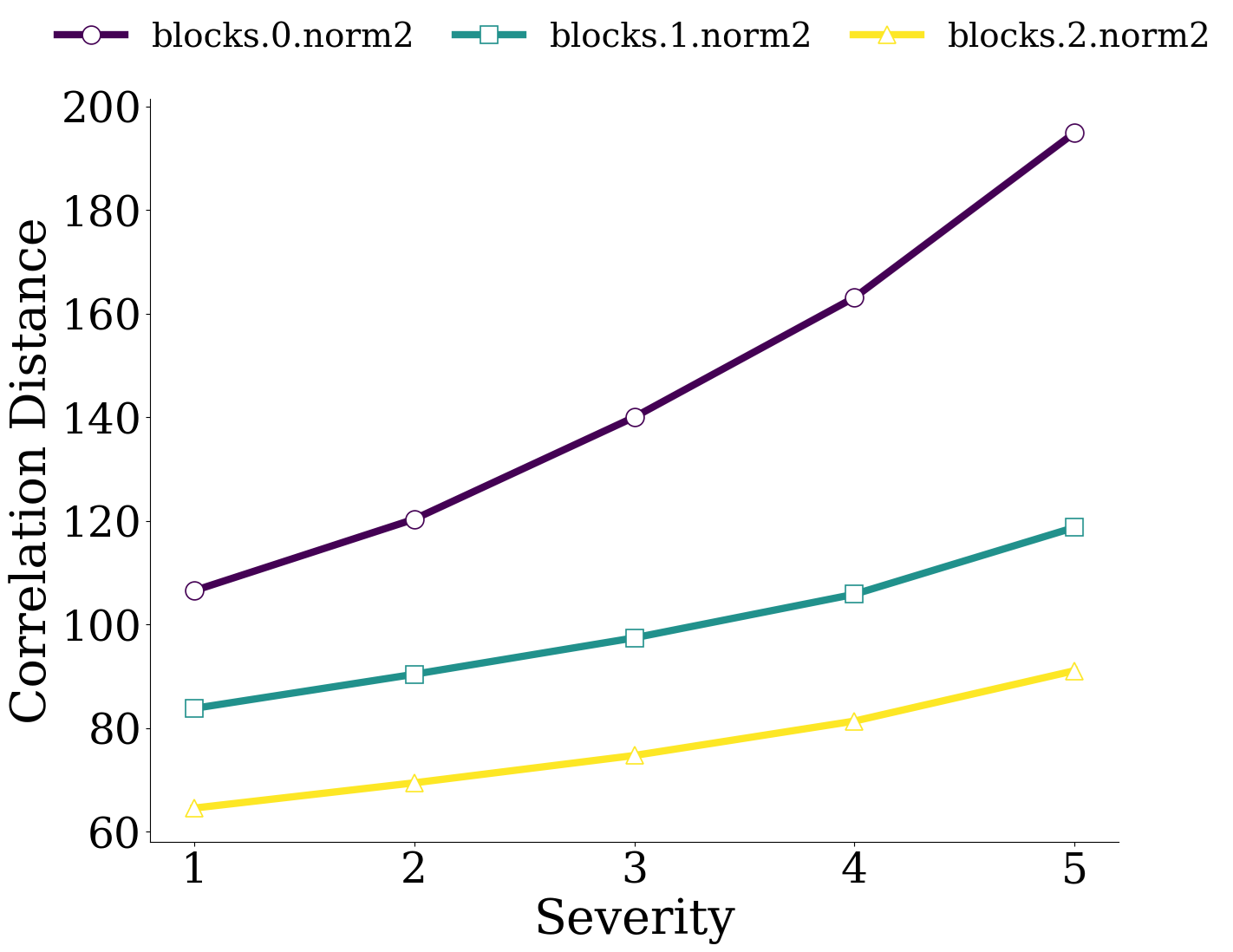}
      \caption{Correlation Distance}
      \label{fig:delta_sigma}
  \end{subfigure}
  \begin{subfigure}[b]{0.25\linewidth}
      \includegraphics[width=\linewidth]{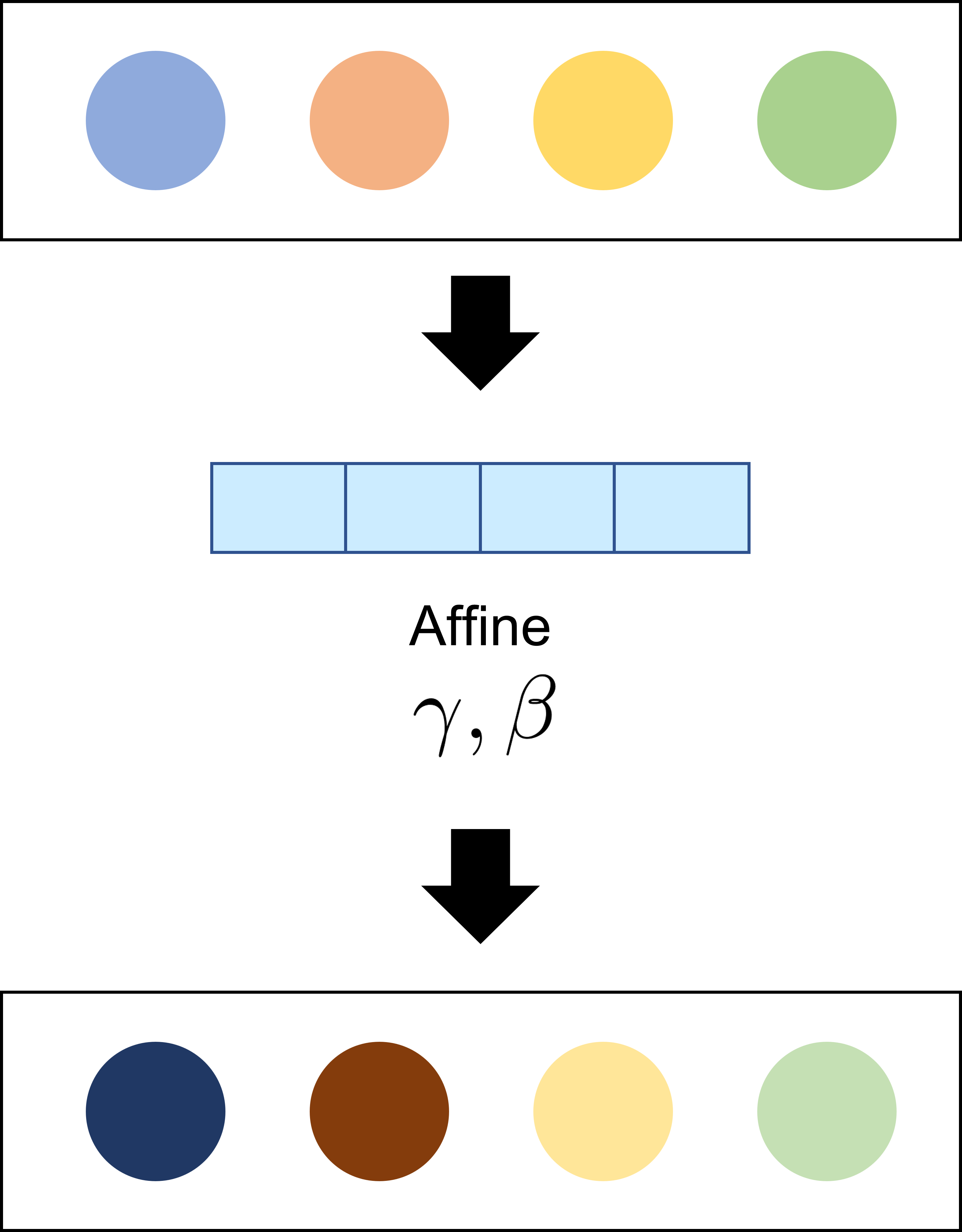}
      \caption{Tent}
      \label{fig:tent}
  \end{subfigure}
  \begin{subfigure}[b]{0.25\linewidth}
      \includegraphics[width=\linewidth]{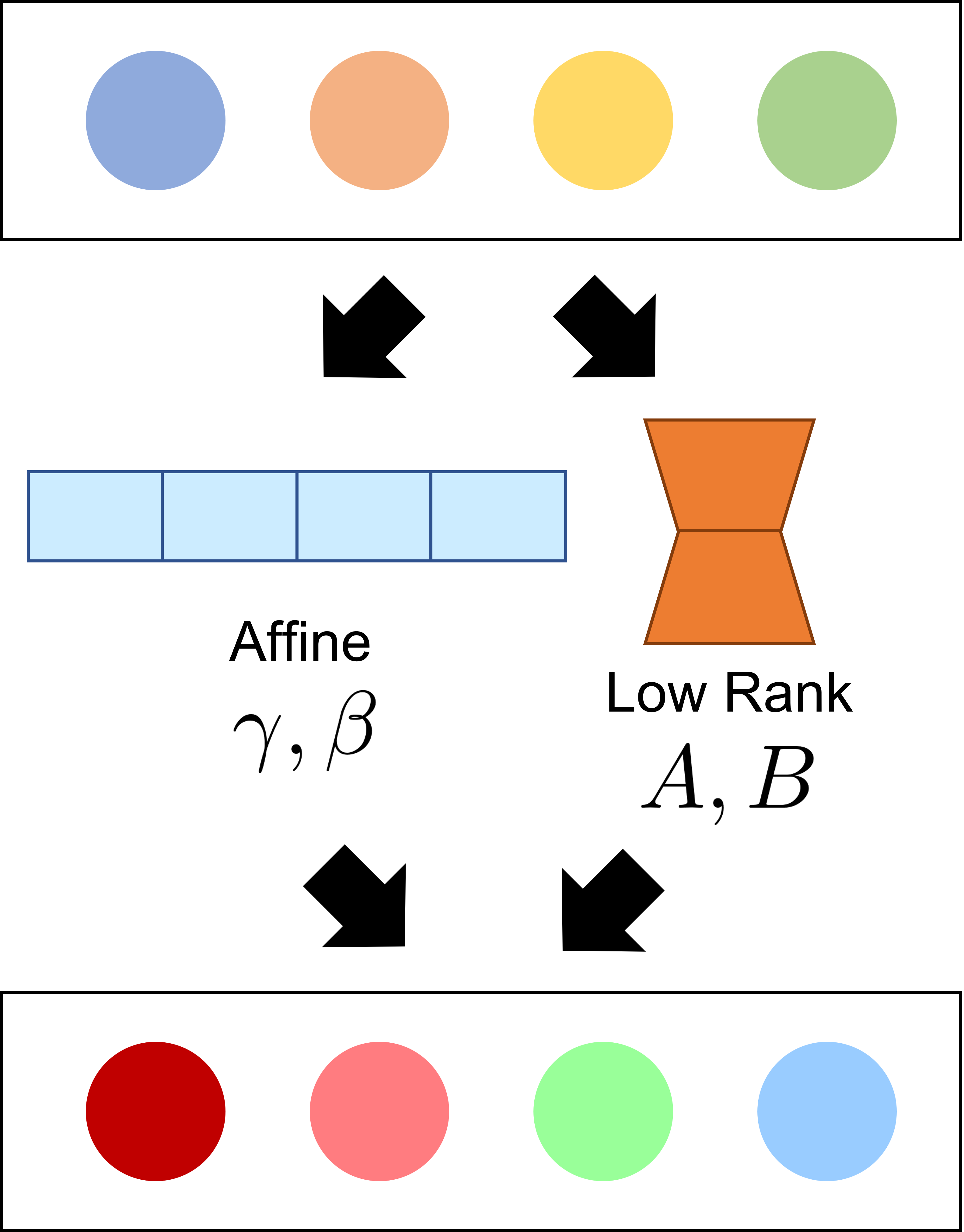}
      \caption{MixTTA (Ours)}
      \label{fig:MixTTA}
  \end{subfigure}
  \caption{(a) Correlation distance comparison across three layers of a pre-trained ViT on ImageNet-C \textit{Gaussian noise} under varying severity levels. (b) and (c) provide overviews of Tent~\cite{wang2020tent} and our proposed MixTTA, respectively, where each colored circle represents a feature dimension. Tent applies only channel-wise affine modulation without mixing across dimensions (colors remain unchanged while intensities vary), whereas MixTTA introduces an additional low-rank cross-channel transform that mixes feature dimensions, leading to changes in both color and intensity.}
\label{fig:overview}
\end{figure*}


In this paper, we revisit this design choice by examining how distribution shift affects intermediate feature representations. 
Specifically, we measure the correlation distance $D = \frac{1}{N}\sum_{n=1}^{N} \|\Sigma^s_n - \Sigma^t_n\|_F$, where $\Sigma^s_n$ and $\Sigma^t_n$ are the per-sample covariance matrices of the source and target features after each normalization layer, $N$ is the number of test samples, and $\|\cdot\|_F$ denotes the Frobenius norm.
As illustrated in Fig. \ref{fig:delta_sigma}, the correlation distance increases progressively with corruption severity across all layers, revealing that distribution shift does not merely alter per-channel variances but fundamentally changes the inter-channel correlation structure. 
Since the learnable affine parameters of normalization layers perform axis-aligned per-channel scaling and shifting, they are \textit{geometrically incapable of correcting such cross-channel structural changes}. 
Moreover, earlier layers exhibit substantially larger correlation distances than deeper layers, indicating that earlier layers are most affected by correlation disruption.
While TCA \cite{you2025tca} proposes aligning feature correlations at test time, it operates only on the final representation, leaving the pronounced early-layer misalignment unaddressed.


Motivated by these observations, we propose \textbf{MixTTA}, a lightweight module that extends normalization layer adaptation to capture cross-channel feature interactions.  As illustrated in Figs. \ref{fig:tent} and \ref{fig:MixTTA}, MixTTA equips the existing per-channel affine transform with a residual cross-channel term parameterized by low-rank matrices, enabling cross-channel mixing at each normalization layer throughout the network. The low-rank design offers a parameter-efficient alternative to full cross-channel mixing, avoiding the instability of unconstrained adaptation while retaining sufficient capacity to capture the dominant correlation shifts. 
To maintain a clear functional separation between the diagonal affine path and the low-rank branch, we propose \textbf{Decoupling Projection} that constrains the low-rank component to purely off-diagonal corrections. Additionally, we further propose \textbf{Spectral Projection} that suppresses rank-1 collapse induced by the dominant principal component, stabilizing adaptation under biased or non-stationary test streams. 

The proposed module operates as a plug-in that can be integrated into any existing normalization-based TTA method that updates axis-aligned per-channel parameters. Extensive experiments across standard and wild TTA scenarios demonstrate that MixTTA achieves state-of-the-art adaptation performance. Notably, under wild scenarios involving class imbalance, single-sample adaptation, and mixed domains, MixTTA substantially mitigates the prediction collapse that commonly afflicts existing methods.

The contributions of our work are summarized as follows:
\begin{itemize}
    \item 
    We show that distribution shift induces substantial changes in inter-channel feature correlations, and that this disruption is most pronounced in earlier layers. This reveals a key limitation of the per-channel affine paradigm in TTA, as per-channel modulation cannot correct cross-channel structural changes.

    \item 
    We introduce MixTTA, a residual low-rank cross-channel transform that extends normalization layer adaptation to model inter-channel dependencies. 
    We additionally propose Decoupling Projection to enforce strict diagonal/off-diagonal separation and Spectral Projection to prevent rank-1 collapse under non-stationary test streams.

    \item
    When integrated into diverse TTA baselines, MixTTA yields consistent accuracy gains across standard and wild scenarios with improved robustness and minimal overhead.

\end{itemize}
\section{Related Work}
\label{sec:related-work}

\subsection{Test-Time Adaptation via Normalization Layers}
Test-time adaptation (TTA) has recently emerged as a promising paradigm for improving model robustness under distribution shifts without requiring access to source data or target labels. 
Tent~\cite{wang2020tent} is a foundational method that minimizes prediction entropy while updating only the affine parameters of normalization layers.
Subsequent works have extended Tent along complementary directions. 
MEMO~\cite{zhang2022memo} augments each test instance and adapts by minimizing the marginal entropy across augmentations.
EATA~\cite{niu2022eata} introduces a Fisher regularizer to preserve critical parameters and selects reliable test samples to mitigate catastrophic forgetting. 
SAR~\cite{niu2023sar} achieves robust online adaptation through batch-agnostic normalization and sharpness-aware optimization.
DeYO~\cite{lee2024entropy} proposes the Pseudo-Label Probability Difference (PLPD) metric, weighting samples by shape consistency when entropy-based confidence is unreliable. 
COME~\cite{zhang2025come} addresses the overconfidence failure mode of entropy minimization by modeling a Dirichlet prior over predictions.
ReCAP~\cite{hu2025beyond} further improves robustness in wild scenarios by leveraging region-level confidence as an adaptation proxy.
Despite their advances, these approaches all operate within Tent's per-channel affine update regime.
Our method, MixTTA, extends this regime by introducing cross-channel mixing through a lightweight low-rank module while maintaining parameter efficiency.

\subsection{Correlation Alignment under Distribution Shift}
Correlation alignment aims to match feature distributions between source and target domains by aligning their covariance structures. 
CORAL \cite{sun2016deepcoral} pioneered this direction by minimizing the distance between second-order statistics, and subsequent works extended it to higher-order moments (HoMM \cite{chen2020homm}, CMD \cite{zellinger2017cmd}) and entropy-aware formulations \cite{morerio2018minimal}.
Recently, TCA \cite{you2025tca} adapted correlation alignment to the TTA scenario by transforming test features to match pseudo-source correlations estimated from high-confidence samples. 
However, TCA operates only on the final feature representation, leaving early-layer correlation misalignment unaddressed. 
In contrast, our approach places cross-channel correction at each normalization layer, enabling hierarchical alignment throughout the network.


\subsection{Low-Rank Adaptation}
Low-rank adaptation (LoRA)~\cite{hu2022lora} factorizes weight updates into low-rank matrices, enabling efficient fine-tuning with minimal additional parameters. 
This paradigm has inspired a range of parameter-efficient adaptation methods such as AdapterFusion~\cite{pfeiffer2021adapterfusion}, Prompt Tuning~\cite{lester2021power}, and Visual Prompt Tuning~\cite{jia2022vpt}. 
Recent efforts have incorporated low-rank modules into test-time adaptation scenarios as well. 
Imam~\etal~\cite{imam2025test} inserted low-rank adapters into transformer attention and optimized them via confidence maximization on unlabeled test data, while Kojima~\etal~\cite{kojima2025lora} performed low-rank test-time training of visual encoders in vision–language models with an auxiliary self-supervised objective. 
Additionally, ViDA~\cite{liu2024vida} targets Continual TTA and injects high-rank and low-rank adapters into linear or convolutional layers, with a Homeostatic Knowledge Allotment strategy to mitigate catastrophic forgetting.
These methods apply low-rank factorization to network weights, whereas MixTTA applies it to inter-channel dependencies within normalization layers, targeting correlation structure rather than weight adaptation.

\section{Methodology}

\subsection{Preliminary}
Given a source model with parameters \(\theta\) and target data \(x\),
Tent~\cite{wang2020tent} performs test-time adaptation by minimizing the prediction entropy:
\begin{equation}\label{eqn:preliminary}
    \mathcal{L}_{\mathrm{Ent}}(x) \;=\; \mathrm{Ent}_{\theta}(x),
\end{equation}
where \(\mathrm{Ent}_{\theta}(x)\) denotes the prediction entropy of the model.
To optimize this objective, Tent updates only the affine parameters of normalization layers 
(\eg, BatchNorm~\cite{ioffe2015batch}, GroupNorm~\cite{wu2018group}, or LayerNorm~\cite{ba2016layer}),
while keeping all other weights frozen to avoid overfitting and catastrophic collapse. 
Formally, Tent modulates the normalized feature \(x \in \mathbb{R}^{C\times T}\) as
\begin{equation}\label{eqn:tent_modulation}
y \;=\; \gamma \odot x  \;+\; \beta,
\end{equation}
where $\odot$ denotes the Hadamard product, \(y \in \mathbb{R}^{C\times T}\) denotes the modulated activation,
\(\gamma,\beta \in \mathbb{R}^{C}\) are learnable per-channel scale and bias parameters,
\(C\) is the number of channels, and \(T\) is the number of tokens 
(\eg, \(T=HW\) for CNNs or sequence length for Vision Transformers). 

\subsection{Low-Rank Cross-Channel Mixing}
Per-channel modulation in Tent and its follow-ups~\cite{niu2022eata, niu2023sar, lee2024entropy} is parameter-efficient and effective, but it cannot explicitly capture cross-channel dependencies since it only rescales each channel independently. 
This limitation is directly reflected in Eq.~\eqref{eqn:tent_modulation}, which can be rewritten in the following matrix form:
\begin{equation}\label{eqn:tent_matrix}
y \;=\; \Gamma x + \beta \mathbf,
\end{equation}
where $\Gamma \in \mathbb{R}^{C \times C}$ is the diagonal matrix with diagonal entries $\gamma$.
This highlights that Tent restricts the channel transform to be diagonal and cannot model off-diagonal cross-channel mixing.
When distribution shifts alter inter-channel correlations, such limited representational capacity can lead to under-adaptation.

\begin{figure}[t]
  \centering
  \includegraphics[width=\linewidth]{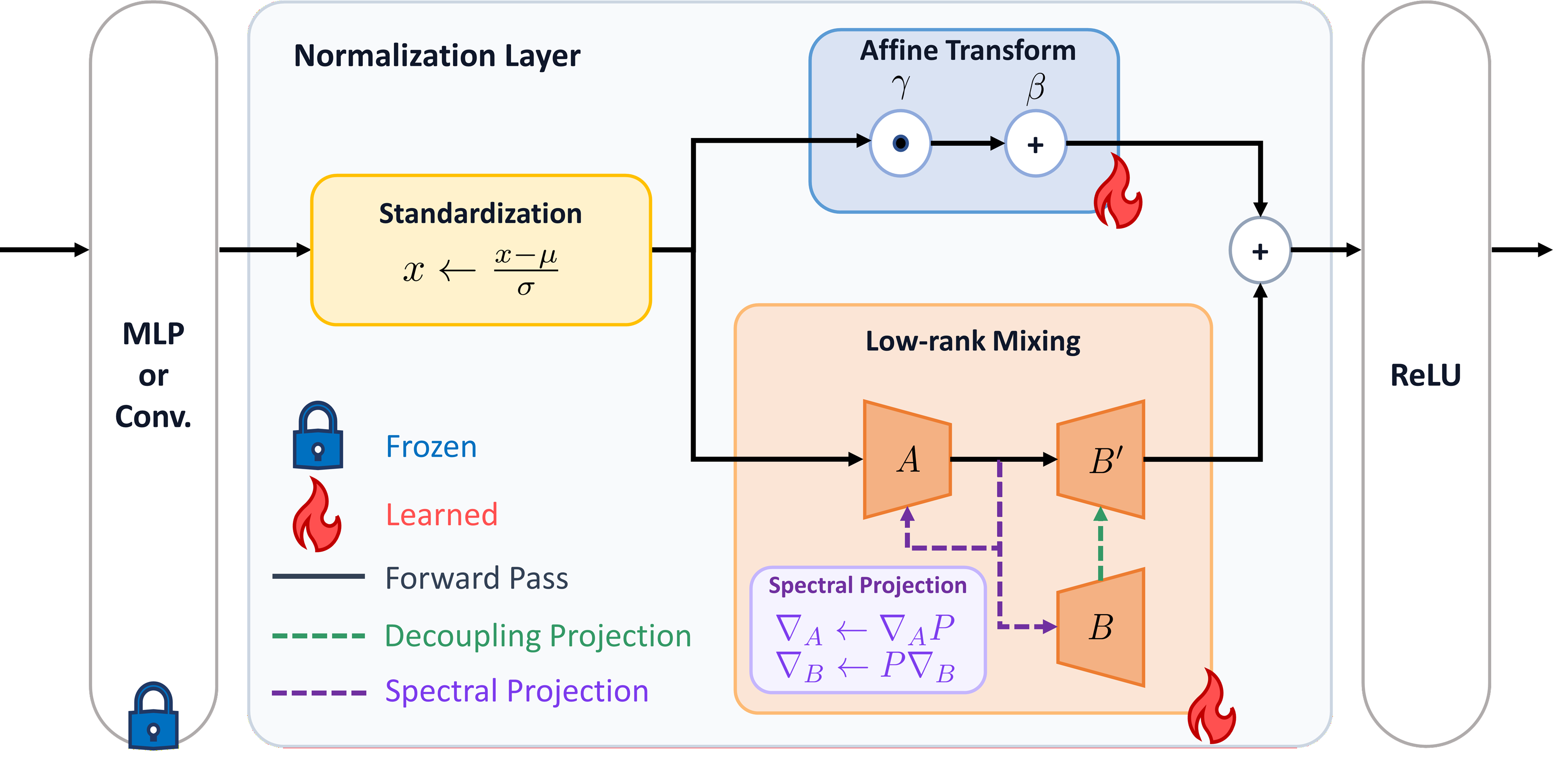}
  \caption{
  Overall structure of MixTTA module, which operates within normalization layers of the backbone network. Standardized features are modulated through both the affine and residual low-rank branches. Decoupling Projection ensures strict separation between diagonal and off-diagonal updates, while Spectral Projection filters collapse-prone directions during adaptation.
  }
  \label{fig:MixTTA Module}
\end{figure}

A naïve remedy for this limitation is to replace $\Gamma$ with a full channel-mixing transform 
\(W \in \mathbb{R}^{C \times C}\) applied after normalization.
Although this modification allows the model to capture inter-channel correlations, 
it introduces \(C^2\) parameters per layer, which substantially increases adaptation cost and makes the model vulnerable to overfitting or collapse during adaptation. 
Moreover, learning a dense $W$ may dilute the influence of the pretrained affine parameters $(\gamma,\beta)$ that normalization layers already provide, removing the strong inductive bias of a diagonal initialization.
To address this, we introduce a \emph{low-rank channel mixing} that extends Tent’s diagonal scaling by a compact rank-\(r\) perturbation:
\begin{equation}\label{eqn:MixTTA_modulation}
    y \;=\; \gamma \odot x \;+\; (AB)^{\top}x \;+\; \beta,
\end{equation}
where \(A \in \mathbb{R}^{C \times r}\) and \(B \in \mathbb{R}^{r \times C}\) are learnable low-rank matrices, and \(r \in \mathbb{N}\) denotes the subspace dimension with \(r \ll C\).
This parameterization decomposes \(W\) as \(W = \Gamma + \Delta\), where $\Delta=( AB)^\top$.
The diagonal term \(\Gamma\) retains Tent’s affine updates, while the low-rank perturbation \(\Delta\) captures cross-channel dependencies by projecting features onto a compact subspace and reconstructing them to the full channel space. 
This formulation yields a unified modulation framework that jointly captures per-channel and cross-channel transformations with only a modest overhead in parameters. In practice, we initialize $B$ as a zero matrix so that $W=\Gamma$ at the start of adaptation, recovering Tent and ensuring stable warm-start behavior. 
Fig. \ref{fig:MixTTA Module} illustrates the overall structure of the proposed MixTTA module.

\subsection{Decoupling Projection}
Although $\Delta$ is intended to capture cross-channel mixing, its non-zero diagonal elements can still induce per-channel modulation and thus duplicate the \emph{functionality} of $\Gamma$.
To enforce a clear functional decoupling of $\Gamma$ and $\Delta$, we impose the constraint
\begin{equation}
\label{eq:diagzero_constraint}
\mathrm{diag}(\Delta)=\mathbf{0},
\end{equation}
where $\operatorname{diag}(\cdot)$ denotes the diagonal vector of a matrix. Noting that $\Delta_{ii}=(AB)_{ii}=a_i^\top b_i$, where $a_i\in\mathbb{R}^{r}$ is the $i$-th row of $A$ and $b_i\in\mathbb{R}^{r}$ is the $i$-th column of $B$, Eq.~\eqref{eq:diagzero_constraint} is satisfied if $a_i^\top b_i=0$ for all $i$.
Accordingly, before the forward pass, we project each $b_i$ onto the orthogonal complement of $a_i$:
\begin{equation}
\label{eq:bproj}
b'_i \leftarrow b_i - \operatorname{sg}\!\left(\frac{a_i^\top b_i}{\|a_i\|_2^2+\epsilon}\, a_i\right),
\qquad \forall i\in\{1,\dots,C\},
\end{equation}
where $\operatorname{sg}(\cdot)$ denotes the stop-gradient operator, $\|\cdot\|_2$ denotes the L2-norm, and $\epsilon>0$ is a small constant for numerical stability.
This projection enforces the diagonal components of $\Delta$ to be zero while preserving the off-diagonal mixing capacity of the low-rank branch.

\subsection{Spectral Projection}
While MixTTA enables the model to exploit cross-channel interactions during test-time adaptation, 
its entropy-driven updates remain susceptible to collapsing into degenerate modes.
In practice, when the target data stream is biased or non-stationary, the adapted model tends to concentrate its optimization along dominant \emph{rank-1} directions.

We now analyze why unconstrained MixTTA can collapse into a rank-1 mode.
Let the upstream gradient $g:=\partial \mathcal{L}_{\mathrm{Ent}}/\partial y\in\mathbb{R}^{C\times T}$.
By standard backpropagation through $B^\top A^\top x$, the gradients of the low-rank factors take the coupled form
\begin{equation}
\label{eq:coupled_grads}
\nabla_A \mathcal{L}_{\mathrm{Ent}} =x \big(Bg\big)^\top,
\qquad
\nabla_B \mathcal{L}_{\mathrm{Ent}} = (A^\top x)\, g^\top.
\end{equation}
Therefore, given step size $\delta$, SGD updates of $(A,B)$ follow an alternating dynamics:
\begin{equation}
\label{eq:power_like_update_noF}
A \leftarrow A - \delta\, x (Bg)^{\!\top},
\qquad
B \leftarrow B - \delta\, (A^{\!\top}x)\, g^{\!\top}.
\end{equation}
Crucially, the update of $A$ depends on $B$ (through $Bg$), and the update of $B$ depends on $A$ (through $A^\top x$),
which creates a positive feedback loop that repeatedly reinforces whichever direction dominates the subspace features.
When the subspace statistics become highly anisotropic (\ie, $\lambda_1\gg\lambda_2$), this alternation tends to align the columns of $A$ and $B$ to the same dominant axis, suppressing cross-channel diversity and effectively degenerating the low-rank update to rank-1.
Empirically, we observe that the top-1 principal component tends to dominate the feature energy and diverge when left unregularized.
To mitigate this excessive rank-1 concentration, we introduce the \textbf{Spectral Projection}.

Given the subspace feature $z := A^{\top}x$, we compute the dominant eigenvector $u_1 \in \mathbb{R}^r$ of the feature covariance as follows:
\begin{equation}
u_1 := \mathrm{eigvec}_{\max}\!\left(\mathrm{Cov}(z)\right),
\end{equation}
where $\mathrm{Cov}(z)$ denotes the covariance matrix of $z$ and $\mathrm{eigvec}_{\max}(\cdot)$ denotes the eigenvector corresponding to the largest eigenvalue, 
which can be efficiently obtained via the power iteration method~\cite{miyato2018spectral}. 
We then construct a projection operator $P$ as follows:
\begin{equation}
\begin{aligned}
    P &:= I - \eta\frac{u_1 u_1^\top}{\|u_1\|_2^2 + \epsilon} \in \mathbb{R}^{r\times r},
\end{aligned}
\end{equation}
where $I\in\mathbb{R}^{r\times r}$ denotes the identity matrix, $\eta \in [0,1]$ controls the projection strength.
During adaptation, the low-rank gradients are projected as
\begin{equation}
    \nabla_A \leftarrow \nabla_A P, 
    \qquad 
    \nabla_B \leftarrow P\nabla_B,
\end{equation}
thereby suppressing the dominant rank-1 direction and stabilizing adaptation.
For batched inputs, we compute a per-sample projection matrix and use their average $\bar{P}$ as the batch-level projection operator. Although $\bar{P}$ is not an exact projector (\ie, $\bar{P}^2 \neq \bar{P}$), we find that it sufficiently suppresses the dominant component and stabilizes adaptation.

\section{Experiments}
\label{sec:Experiments}
\definecolor{lightgray}{gray}{0.9}
\newcommand{\meanpm}[2]{#1{\scriptsize$\pm\,#2$}}
\newcommand{\cstat}[2]{\makecell{#1\\[-2pt]{\scriptsize$\pm$#2}}} 

\subsection{Benchmarks, Baselines, and Test Scenarios}
We evaluated our method on ImageNet-C~\cite{hendrycks2019benchmarking} and ImageNet-Sketch~\cite{wang2019learning}, two standard benchmarks for TTA. 
ImageNet-C is derived from the original ImageNet dataset~\cite{russakovsky2015imagenet} and designed to assess model robustness under a wide range of common corruptions. 
It includes 15 corruption types each applied at five severity levels. 
ImageNet-Sketch, on the other hand, contains sketch-based depictions of the 1000 ImageNet categories, providing substantial appearance-level deviations from natural images. 
This benchmark focuses on cross-domain generalization rather than simple corruption.
For adaptation strategies, we evaluated MixTTA by integrating it into representative TTA baselines, including Tent~\cite{wang2020tent}, EATA~\cite{niu2022eata}, SAR~\cite{niu2023sar}, DeYO~\cite{lee2024entropy}, and ReCAP~\cite{hu2025beyond}. 
We additionally included LinearTCA and LinearTCA$^+$ ~\cite{you2025tca} as a baseline that addresses cross-channel correlation at test time.
Following the convention in \cite{you2025tca}, where LinearTCA$^{+}$ denotes LinearTCA applied to the best-performing baseline, we instantiated LinearTCA$^{+}$ using ReCAP (\ie, ReCAP+LinearTCA) to enable a fair and consistent comparison with a state-of-the-art base method.
Finally, we evaluated under the mild and three wild test scenarios proposed by Niu et al.~\cite{niu2023sar}. Wild scenarios include (i) \textit{online imbalanced label shift}, which models fluctuations in the ground-truth label distribution,
(ii) \textit{single-sample} setting (\textit{i.e.}, batch size 1), which evaluates a model’s ability to adapt without relying on batch statistics, and
(iii)  \textit{mixed shift}, which combines multiple types of distribution shifts simultaneously.

\subsection{Implementation Details}
Unless otherwise noted, we followed the experimental protocol of~\cite{lee2024entropy}. 
All experiments were repeated with three random seeds and we report their mean.
We employed ViT~\cite{dosovitskiy2020image} with Layer Normalization ~\cite{ba2016layer} as backbone models and pre-trained weights were obtained from the \texttt{timm} library~\cite{rw2019timm}. 
The learning rate was set to 0.001 and it was doubled under the batch size 1 scenario to compensate for the reduced reliability of batch-level statistics. 
We initialized the matrix $A$ using the Xavier scheme~\cite{glorot2010understanding} and set $B$ to the zero matrix.
MixTTA module was applied to the second normalization layer in each of the first five transformer encoder blocks.
The subspace dimension $r$ and spectral projection strength $\eta$ were fixed to 4 and 0.9 across all experiments, respectively. 
Due to the page limit, full ResNet results are provided in the supplementary material; MixTTA consistently improves most baselines on ImageNet-C and ImageNet-Sketch across all four adaptation scenarios.

\begin{table*}[t]
\centering
\small
\setlength{\tabcolsep}{1mm}
\resizebox{\textwidth}{!}{%
\begin{tabular}{l|ccc|cccc|cccc|cccc|c}
\multirow{2}{*}{\textbf{Mild}} & \multicolumn{3}{c|}{\textbf{Noise}} & \multicolumn{4}{c|}{\textbf{Blur}} & \multicolumn{4}{c|}{\textbf{Weather}} & \multicolumn{4}{c|}{\textbf{Digital}} & \multirow{2}{*}{\textbf{Avg.}} \\
\cline{2-16}
& Gauss. & Shot & Impulse & Defocus & Glass & Motion & Zoom & Snow & Frost & Fog & Brit. & Contr. & Elastic & Pixel & JPEG & \\
\hline
No Adapt. & 16.8 & 12.0 & 16.5 & 29.2 & 23.6 & 34.0 & 27.3 & 15.8 & 26.6 & 47.5 & 55.4 & 44.3 & 31.0 & 45.1 & 49.1 & 31.6 \\
\hline
$\bullet$ LinearTCA    & 17.9 & 13.0 & 17.4 & 29.8 & 24.2 & 35.7 & 28.8 & 17.0 & 27.8 & 49.8 & 56.0 & 45.8 & 33.0 & 46.0 & 49.8 & 32.8 $\pm$ 0.0 \\
\hline
$\bullet$ LinearTCA$^+$    & 54.3 & 54.6 & 55.4 & 58.3 & 58.4 & 63.2 & 60.1 & 66.8 & 65.7 & 73.4 & 78.1 & 68.3 & 67.4 & 73.1 & 70.2 & 64.5 $\pm$ 0.3 \\
\hline
$\bullet$ Tent         & 44.6 & \textbf{42.8} & \textbf{45.2} & \textbf{52.1} & 47.8 & 55.3 & 50.3 & 18.1 & 21.2 & 66.4 & 75.0 & 64.8 & 52.7 & 66.9 & 64.4 & 51.2 $\pm$ 0.0\\
\rowcolor{lightgray}
~~~+MixTTA  & \textbf{47.9} & 38.7 & 41.3 & 51.3 & \textbf{48.3} & \textbf{56.2} & \textbf{51.0} & \textbf{57.2} & \textbf{41.6} & \textbf{68.0} & \textbf{75.3} & \textbf{64.9} & \textbf{53.6} & \textbf{68.2} & \textbf{65.1} & \textbf{55.2} $\pm$ 1.4 \\
\hline
$\bullet$ EATA         & 51.7 & 51.6 & 52.4 & 55.5 & 55.8 & 60.2 & 57.7 & 63.1 & 61.0 & 71.2 & 75.4 & 67.1 & 64.2 & 70.5 & 67.9 & 61.7 $\pm$ 0.3\\
\rowcolor{lightgray}
~~~+MixTTA  & \textbf{53.9} & \textbf{53.5} & \textbf{55.0} & \textbf{56.3} & \textbf{57.4} & \textbf{62.1} & \textbf{59.7} & \textbf{66.5} & \textbf{64.6} & \textbf{72.4} & \textbf{77.5} & \textbf{67.6} & \textbf{65.1} & \textbf{72.7} & \textbf{68.8} & \textbf{63.5} $\pm$ 0.1\\
\hline
$\bullet$ SAR          & 46.3 & 45.0 & 46.9 & \textbf{52.9} & \textbf{50.0} & \textbf{55.9} & \textbf{51.5} & 57.0 & 53.5 & 66.5 & 74.8 & \textbf{64.4} & \textbf{55.2} & 66.8 & \textbf{64.5} & 56.7 $\pm$ 0.2\\
\rowcolor{lightgray}
~~~+MixTTA   & \textbf{47.8} & \textbf{46.9} & \textbf{49.1} & 51.7 & 49.0 & 55.7 & 50.4 & \textbf{58.9} & \textbf{57.4} & \textbf{67.0} & 74.8 & 64.1 & 53.9 & \textbf{67.4} & 64.2 & \textbf{57.2} $\pm$ 0.3\\
\hline
$\bullet$ DeYO         & 54.7 & 55.1 & 55.7 & \textbf{58.4} & 58.9 & 63.4 & 45.8 & 67.2 & 65.8 & 73.3 & 78.3 & 68.0 & 68.0 & 73.4 & 70.4 & 63.8 $\pm$ 0.6\\
\hline
\rowcolor{lightgray}
~~~+MixTTA  & \textbf{56.0} & \textbf{56.6} & \textbf{57.0} & 58.2 & \textbf{59.3} & \textbf{64.6} & \textbf{62.2} & \textbf{68.6} & \textbf{67.1} & \textbf{74.0} & \textbf{78.5} & \textbf{68.6} & \textbf{68.4} & \textbf{74.2} & \textbf{71.0} & \textbf{65.6} $\pm$ 0.1\\
\hline
$\bullet$ ReCAP        & 53.9 & 54.2 & 55.0 & \textbf{57.9} & 58.1 & 62.6 & 58.9 & 66.4 & 65.1 & 72.9 & 78.0 & 67.9 & 67.0 & 72.7 & 69.9 & 64.0 $\pm$ 0.1\\
\rowcolor{lightgray}
~~~+MixTTA & \textbf{55.4} & \textbf{55.7} & \textbf{56.4} & 57.6 & \textbf{58.6} & \textbf{63.6} & \textbf{60.9} & \textbf{67.9} & \textbf{66.4} & \textbf{73.4} & \textbf{78.2} & \textbf{68.4} & \textbf{67.1} & \textbf{73.5} & \textbf{70.4} & \textbf{64.9} $\pm$ 0.1\\
\end{tabular}
}
\caption{
Comparisons with baselines on ImageNet-C at severity level 5 under the mild scenario regarding accuracy (\%). \textbf{Bold} values denote the top-performing results.}
\label{tab:comparison mild}
\end{table*}

\subsection{Main Results}
\subsubsection{Comparison on Mild Scenario}
Table~\ref{tab:comparison mild} presents the results on the ImageNet-C dataset at severity level~5 under the mild scenario. 
Integrating MixTTA improves the average accuracy across all five baselines, confirming its generality as a plug-in module.
In particular, for ReCAP, MixTTA delivers a larger improvement (+0.9\%p) in average accuracy than applying LinearTCA (+0.5\%p).
Furthermore, the gains are especially pronounced on challenging \textit{Noise} and \textit{Weather} corruptions. For example, MixTTA improves Tent from 18.1\% to 57.2\% on \textit{Snow} and from 21.2\% to 41.6\% on \textit{Frost}. 
These results suggest that MixTTA is particularly beneficial under corruptions where baseline adaptation is most fragile.

\begin{table*}[t]
\centering
\small
\setlength{\tabcolsep}{1mm}
\resizebox{\textwidth}{!}{%
\begin{tabular}{l|ccc|cccc|cccc|cccc|c}
\multirow{2}{*}{\textbf{Label Shifts}} & \multicolumn{3}{c|}{\textbf{Noise}} & \multicolumn{4}{c|}{\textbf{Blur}} & \multicolumn{4}{c|}{\textbf{Weather}} & \multicolumn{4}{c|}{\textbf{Digital}} & \multirow{2}{*}{\textbf{Avg.}} \\
\cline{2-16}
& Gauss. & Shot & Impulse & Defocus & Glass & Motion & Zoom & Snow & Frost & Fog & Brit. & Contr. & Elastic & Pixel & JPEG & \\
\hline
No Adapt. & 16.7 & 11.9 & 16.3 & 29.3 & 23.6 & 33.9 & 27.3 & 15.9 & 26.5 & 47.2 & 54.9 & 44.1 & 30.9 & 45.0 & 48.9 & 31.5 \\
\hline
$\bullet$ LinearTCA    & 18.0 & 13.0 & 17.4 & 30.1 & 24.4 & 36.0 & 29.0 & 17.2 & 28.0 & 50.0 & 55.6 & 45.9 & 33.1 & 46.2 & 49.8 & 32.9 $\pm$ 0.0 \\
\hline
$\bullet$ LinearTCA$^+$    & 55.1 & 55.1 & 56.0 & 58.3 & 58.9 & 63.8 & 61.9 & 67.9 & 66.3 & 73.3 & 78.0 & 67.1 & 69.0 & 73.5 & 70.4 & 65.0 $\pm$ 0.1  \\
\hline
$\bullet$ Tent         & \textbf{46.4} & \textbf{46.8} & \textbf{46.1} & \textbf{54.6} & 52.2 & 58.2 & 53.0 & 10.2 & 10.5 & 69.6 & 76.2 & 66.1 & 58.1 & 69.4 & 66.7 & 52.3 $\pm$ 0.3 \\
\rowcolor{lightgray}
~~~+MixTTA   & 37.8 & 34.0 & 29.4 & 54.2 & \textbf{53.6} & \textbf{59.6} & \textbf{55.1} & \textbf{45.5} & \textbf{21.0} & \textbf{70.9} & \textbf{76.5} & \textbf{66.7} & \textbf{60.1} & \textbf{71.0} & \textbf{68.1} & \textbf{53.6} $\pm$ 2.5  \\
\hline
$\bullet$ EATA         & 39.3 & 37.6 & 38.9 & 44.1 & 45.2 & \textbf{44.4} & 46.8 & 54.7 & 54.1 & \textbf{60.0} & 72.5 & \textbf{25.9} & 58.0 & 66.3 & 64.0 & 50.1 $\pm$ 0.7  \\
\rowcolor{lightgray}
~~~+MixTTA   & \textbf{47.9} & \textbf{46.2} & \textbf{48.9} & \textbf{50.3} & \textbf{52.6} & 31.6 & \textbf{54.2} & \textbf{62.9} & \textbf{60.6} & 58.7 & \textbf{76.0} & 19.1 & \textbf{62.6} & \textbf{71.0} & \textbf{66.1} & \textbf{53.9} $\pm$ 3.3  \\
\hline
$\bullet$ SAR          & 50.0 & 49.1 & 50.6 & \textbf{55.4} & 54.1 & 59.1 & 54.6 & 56.9 & 48.6 & 69.7 & 76.3 & 66.2 & \textbf{60.8} & 69.7 & 66.9 & 59.2 $\pm$ 0.6  \\
\rowcolor{lightgray}
~~~+MixTTA   & \textbf{52.0} & \textbf{51.2} & \textbf{52.7} & 54.6 & \textbf{54.6} & \textbf{59.5} & \textbf{54.8} & \textbf{63.1} & \textbf{58.4} & \textbf{70.4} & 76.3 & \textbf{66.3} & 60.1 & \textbf{70.5} & \textbf{67.5} & \textbf{60.8} $\pm$ 0.2  \\
\hline
$\bullet$ DeYO         & 54.5 & 55.1 & 55.8 & \textbf{58.0} & 58.9 & 63.8 & 61.4 & 67.9 & 66.2 & 73.1 & 78.0 & 66.7 & 69.1 & 73.6 & 70.5 & 64.8 $\pm$ 0.1 \\
\rowcolor{lightgray}
~~~+MixTTA   & \textbf{56.0} & \textbf{56.3} & \textbf{56.9} & 57.9 & \textbf{59.3} & \textbf{64.7} & \textbf{62.8} & \textbf{68.9} & \textbf{67.2} & \textbf{73.9} & \textbf{78.4} & \textbf{67.4} & \textbf{69.2} & \textbf{74.2} & \textbf{71.1} & \textbf{65.6} $\pm$ 0.1 \\
\hline
$\bullet$ ReCAP        & 54.4 & 54.9 & 55.6 & 58.1 & 58.9 & 63.6 & 46.9 & 67.8 & 66.1 & 73.1 & 77.9 & 66.8 & 68.7 & 73.4 & 70.4 & 63.8 $\pm$ 1.7 \\
\rowcolor{lightgray}
~~~+MixTTA   & \textbf{55.8} & \textbf{56.1} & \textbf{56.8} & 57.8 & \textbf{59.3} & \textbf{64.9} & \textbf{63.5} & \textbf{69.0} & \textbf{67.2} & \textbf{73.8} & \textbf{78.4} & \textbf{67.9} & \textbf{69.3} & \textbf{74.3} & \textbf{71.0} & \textbf{65.7} $\pm$ 0.1 \\
\hline

\end{tabular}
}
\caption{
Comparisons with baselines on ImageNet-C at severity level 5 under online imbalanced label shifts with imbalance ratio $\infty$ regarding accuracy (\%).}
\label{tab:comparison label shifts}
\end{table*}

\begin{table*}[t]
\centering
\small
\setlength{\tabcolsep}{1mm}
\resizebox{\textwidth}{!}{%
\begin{tabular}{l|ccc|cccc|cccc|cccc|c}
\multirow{2}{*}{\textbf{Batch Size 1}} & \multicolumn{3}{c|}{\textbf{Noise}} & \multicolumn{4}{c|}{\textbf{Blur}} & \multicolumn{4}{c|}{\textbf{Weather}} & \multicolumn{4}{c|}{\textbf{Digital}} & \multirow{2}{*}{\textbf{Avg.}} \\
\cline{2-16}
& Gauss. & Shot & Impulse & Defocus & Glass & Motion & Zoom & Snow & Frost & Fog & Brit. & Contr. & Elastic & Pixel & JPEG & \\
\hline
No Adapt. & 16.8 & 12.0 & 16.5 & 29.2 & 23.6 & 34.0 & 27.3 & 15.8 & 26.6 & 47.5 & 55.4 & 44.3 & 31.0 & 45.1 & 49.1 & 31.6 \\
\hline
$\bullet$ LinearTCA    & 17.9 & 13.0 & 17.4 & 29.8 & 24.2 & 35.7 & 28.8 & 17.0 & 27.8 & 49.8 & 56.0 & 45.8 & 33.0 & 46.0 & 49.8 & 32.8 $\pm$ 0.0 \\
\hline
$\bullet$ LinearTCA$^+$    & 56.1 & 56.9 & 57.5 & 59.1 & 60.4 & 65.6 & 61.1 & 69.3 & 67.5 & 74.0 & 78.6 & 68.0 & 70.3 & 74.6 & 71.5 & 66.0 $\pm$ 0.0  \\
\hline
$\bullet$ Tent         & \textbf{44.7} & \textbf{42.9} & \textbf{45.4} & \textbf{52.5} & 48.1 & 55.5 & 50.6 & 15.6 & 18.6 & 66.6 & 75.1 & 64.9 & 52.9 & 67.1 & 64.6 & 51.0 $\pm$ 0.1 \\
\rowcolor{lightgray}
~~~+MixTTA   & 41.7 & 36.7 & 41.0 & 51.8 & \textbf{48.8} & \textbf{56.4} & \textbf{51.3} & \textbf{57.9} & \textbf{38.8} & \textbf{68.3} & \textbf{75.4} & \textbf{65.2} & \textbf{54.1} & \textbf{68.4} & \textbf{65.4} & \textbf{54.8} $\pm$ 2.5 \\
\hline
$\bullet$ EATA         & 36.4 & 30.7 & 35.4 & 44.7 & 40.0 & 46.4 & 41.8 & 38.5 & 39.1 & 61.8 & 65.9 & 61.6 & 47.1 & 59.7 & 59.7 & 47.2 $\pm$ 0.5 \\
\rowcolor{lightgray}
~~~+MixTTA   & \textbf{49.7} & \textbf{47.6} & \textbf{51.0} & \textbf{55.2} & \textbf{54.4} & \textbf{60.5} & \textbf{56.3} & \textbf{63.8} & \textbf{62.3} & \textbf{71.5} & \textbf{77.1} & \textbf{66.7} & \textbf{61.8} & \textbf{72.0} & \textbf{68.4} & \textbf{61.2} $\pm$ 0.3 \\
\hline
$\bullet$ SAR          & 45.6 & 43.2 & 46.3 & \textbf{53.6} & \textbf{50.5} & \textbf{57.6} & \textbf{53.0} & 58.7 & 54.9 & 68.9 & 75.4 & \textbf{65.7} & \textbf{58.1} & \textbf{69.0} & \textbf{66.4} & 57.8 $\pm$ 0.3 \\
\rowcolor{lightgray}
~~~+MixTTA   & \textbf{47.1} & \textbf{44.9} & \textbf{48.4} & 52.0 & 49.5 & 57.0 & 51.6 & \textbf{59.5} & \textbf{58.7} & \textbf{69.0} & \textbf{76.0} & 65.4 & 56.0 & 67.4 & 66.1 & \textbf{57.9} $\pm$ 0.2 \\
\hline
$\bullet$ DeYO         & 55.2 & 55.6 & 56.3 & \textbf{58.9} & 59.4 & 64.3 & 44.2 & 68.1 & 66.5 & 73.8 & 78.4 & 68.3 & 69.0 & 73.9 & 70.9 & 64.2 $\pm$ 0.6 \\
\rowcolor{lightgray}
~~~+MixTTA   & \textbf{56.2} & \textbf{56.8} & \textbf{57.2} & 58.6 & \textbf{59.9} & \textbf{65.6} & \textbf{63.9} & \textbf{69.3} & \textbf{67.6} & \textbf{74.5} & \textbf{78.8} & \textbf{69.0} & \textbf{69.4} & \textbf{74.7} & \textbf{71.5} & \textbf{66.2} $\pm$ 0.1 \\
\hline
$\bullet$ ReCAP        & 56.1 & 56.8 & 57.2 & \textbf{59.2} & 60.0 & 65.5 & 57.0 & 69.2 & 67.3 & 74.0 & 78.5 & 67.8 & 70.1 & 74.4 & 71.3 & 65.6 $\pm$ 0.8 \\
\rowcolor{lightgray}
~~~+MixTTA   & \textbf{56.6} & \textbf{57.5} & \textbf{57.9} & 58.6 & \textbf{60.6} & \textbf{66.3} & \textbf{66.5} & \textbf{70.1} & \textbf{68.3} & \textbf{74.5} & \textbf{78.8} & \textbf{67.9} & \textbf{70.8} & \textbf{75.0} & \textbf{71.8} & \textbf{66.8} $\pm$ 0.1 \\
\hline
\end{tabular}
}
\caption{
Comparisons with baselines on ImageNet-C at severity level 5 under batch size 1 regarding accuracy (\%).
}
\label{tab:comparison batch size 1}
\end{table*}

\subsubsection{Comparison on Wild Scenario}
Table~\ref{tab:comparison label shifts} reports results under the online imbalanced label shift setting (imbalance ratio $\infty$) on ImageNet-C at severity level~5.
Across a wide range of corruptions, MixTTA consistently improves existing TTA methods, with ReCAP increasing from 63.8\% to 65.7\% and DeYO from 64.8\% to 65.6\% on average.
Beyond the average gains, MixTTA is particularly beneficial in difficult regimes where vanilla adaptation becomes fragile. 
For example, Tent on \textit{Snow} drops accuracy from 15.9\% (No Adapt.) to 10.2\%, but combining it with MixTTA recovers and boosts the accuracy to 45.5\%.
These results indicate that our proposed MixTTA can effectively counteract drift and instability induced by heavily skewed target streams.

Table~\ref{tab:comparison batch size 1} summarizes the ImageNet-C results at severity level~5 in the batch size 1 setting, where adaptation must rely solely on instance-level statistics. 
MixTTA consistently improves strong baselines, boosting the average accuracy of DeYO to 66.2\% and ReCAP to 66.8\%. 
Notably, the gains are particularly evident for methods that degrade under single-sample updates (\eg, EATA improves from 47.2\% to 61.2\% on average), and MixTTA yields broad improvements across most corruption categories, including several \textit{Weather} and \textit{Digital} corruptions. 
Overall, these results indicate that MixTTA remains effective even under highly constrained adaptation, improving robustness while maintaining stable adaptation dynamics.

The results under the mixed-shift setting are summarized in Table~\ref{tab:comparison_mix}. 
MixTTA consistently strengthens adaptation performance across all baselines. 
Notably, while LinearTCA$^+$ yields only a marginal improvement (+0.2\%p) in this setting, MixTTA provides substantially larger gains (+2.8\%p for DeYO and +2.3\%p for ReCAP), highlighting the benefit of its cross-channel correction under mixed shifts.
Taken together, these observations suggest that MixTTA is particularly effective when multiple types of distribution shift co-occur and per-channel adaptation alone is most insufficient. 

\begin{table}[!t]
\centering

\begin{minipage}[t]{0.48\linewidth}
\centering
\small
\scalebox{0.76}{%
\begin{tabular}{l|c}
\rowcolor{white}
\textbf{Mixed Shifts} & \textbf{Accuracy (\%)} \\
\hline
No adapt. & 31.6 \\
\hline
$\bullet$ LinearTCA & 32.8 $\pm$ 0.1  \\
\hline
$\bullet$ LinearTCA$^+$ & 60.3 $\pm$ 0.2 \\
\hline
$\bullet$ Tent & 16.2 $\pm$ 2.5 \\
\rowcolor{lightgray}
~~~+ MixTTA  & \textbf{26.9} $\pm$ 2.7 \\
\hline
$\bullet$ EATA & 56.3 $\pm$ 0.4 \\
\rowcolor{lightgray}
~~~+ MixTTA  & \textbf{61.1} $\pm$ 0.1  \\
\hline
$\bullet$ SAR & 57.7 $\pm$ 0.0 \\
\rowcolor{lightgray}
~~~+ MixTTA  & \textbf{58.9} $\pm$ 0.1 \\
\hline
$\bullet$ DeYO & 59.7 $\pm$ 0.1 \\
\rowcolor{lightgray}
~~~+ MixTTA  & \textbf{62.5} $\pm$ 0.1 \\
\hline
$\bullet$ ReCAP & 60.1 $\pm$ 0.0 \\
\rowcolor{lightgray}
~~~+ MixTTA  & \textbf{62.4} $\pm$ 0.1 \\
\end{tabular}
}
\caption{Comparisons with baselines on ImageNet-C at severity level 5 under a mixture of 15 types of corruption regarding accuracy (\%).}
\label{tab:comparison_mix}
\end{minipage}
\hfill
\begin{minipage}[t]{0.50\linewidth}
\centering
\small
\setlength{\tabcolsep}{1mm}
\resizebox{\linewidth}{!}{%
\begin{tabular}{l|c|c|c}
\rowcolor{white}
{\makecell{\textbf{ImageNet}\\{\textbf{-Sketch}}}} & \textbf{Mild} & \textbf{Label Shifts} & \textbf{Batch Size 1} \\
\hline
No Adapt. & 18.2 & 18.3 & 18.2 \\
\hline
$\bullet$ LinearTCA & 19.3 $\pm$ 0.0  & 19.6 $\pm$ 0.0  & 19.3 $\pm$ 0.0  \\
\hline
$\bullet$ LinearTCA$^+$ & 42.2 $\pm$ 0.2 & 44.6 $\pm$ 0.4 & 43.8 $\pm$ 0.1 \\ 
\hline
$\bullet$ Tent & 8.8 $\pm$ 0.4 & 5.3 $\pm$ 1.4 & 6.1 $\pm$ 0.8 \\
\rowcolor{lightgray}
~~~+ MixTTA & \textbf{34.3} $\pm$ 0.2 & \textbf{29.3} $\pm$ 11.6 & \textbf{34.1} $\pm$ 0.3\\
\hline
$\bullet$ EATA & 36.9 $\pm$ 0.1 & 33.5 $\pm$ 0.0 & 21.5 $\pm$ 0.1 \\
\rowcolor{lightgray}
~~~+ MixTTA & \textbf{41.3} $\pm$ 0.3 & \textbf{37.2} $\pm$ 0.2 & \textbf{36.2} $\pm$ 0.5 \\
\hline
$\bullet$ SAR & 26.3 $\pm$ 2.9 & 13.6 $\pm$ 8.8 & 18.3 $\pm$ 6.3 \\
\rowcolor{lightgray}
~~~+ MixTTA & \textbf{35.6} $\pm$ 0.1 & \textbf{39.3} $\pm$ 0.1 & \textbf{35.5} $\pm$ 0.0 \\
\hline
$\bullet$ DeYO & 43.6 $\pm$ 0.3 & 43.2 $\pm$ 1.0 & 43.4 $\pm$ 0.8 \\
\rowcolor{lightgray}
~~~+ MixTTA & \textbf{43.7} $\pm$ 0.1 & \textbf{44.5} $\pm$ 0.3 & \textbf{44.0} $\pm$ 0.0 \\
\hline
$\bullet$ ReCAP & 42.1 $\pm$ 0.1 & 44.4 $\pm$ 0.4 & 42.9 $\pm$ 0.1 \\
\rowcolor{lightgray}
~~~+ MixTTA & \textbf{42.9} $\pm$ 0.0 & \textbf{45.1} $\pm$ 0.1 & \textbf{45.9} $\pm$ 0.1 \\
\end{tabular}
}
\caption{Comparisons with baselines on ImageNet-Sketch under mild and wild scenarios regarding accuracy (\%).}
\label{tab:comparison_imagenet_sketch}
\end{minipage}

\end{table}

\subsubsection{Results on ImageNet-Sketch}
Table~\ref{tab:comparison_imagenet_sketch} reports accuracy on ImageNet-Sketch under mild and wild settings. 
MixTTA consistently improves existing TTA baselines, with gains becoming markedly larger in wild regimes. 
The most pronounced effect is on Tent, where MixTTA recovers severely degraded performance by at least 24 percentage points across all three settings, indicating that it effectively stabilizes fragile entropy-minimization updates. 
Moreover, ReCAP with MixTTA outperforms LinearTCA$^+$ across all settings, demonstrating that MixTTA provides complementary benefits beyond explicit correlation alignment. 
These results suggest that the cross-channel correction provided by MixTTA is not limited to corruption-based shifts but extends to style-level domain gaps.

\begin{table}[t]
\centering
\small
\setlength{\tabcolsep}{2mm}
\resizebox{0.85\linewidth}{!}{%
\begin{tabular}{r|c|c|c|c|c|c}
\multirow{2}{*}{\textbf{Method}} & \multicolumn{5}{c|}{$r$} & \multirow{2}{*}{Full rank} \\
\cline{2-6}
 & 1 & 2 & 4 & 8 & 16 & \\
\hline
Tent+MixTTA & 53.7 & 54.7 & \textbf{55.2} & 55.0 & 54.6 & 0.1 \\
EATA+MixTTA & 62.4 & 62.5 & \textbf{63.5} & 62.7 & 62.5 & 0.1 \\
SAR+MixTTA & 55.0 & 55.9 & \textbf{57.2} & 56.6 & 55.9 & 0.1 \\
DeYO+MixTTA & 64.0 & 65.1 & \textbf{65.6} & 65.4 & 64.9 & 0.1 \\
ReCAP+MixTTA & 63.6 & 64.2 & \textbf{64.9} & 64.1 & 63.6 & 0.1 \\
\end{tabular}%
}
\caption{
Comparison of average accuracy of MixTTA performance with varying $r$ values and baselines on ImageNet-C severity level 5. Full rank denotes replacing $\Gamma$ with a full channel mixing transform $W \in \mathbb{R}^{C\times C}.$
}
\label{tab:sensitivity_r}
\end{table}

\subsection{Analysis}
\subsubsection{Sensitivity of Subspace Dimension $r$.}
Table~\ref{tab:sensitivity_r} studies the sensitivity of baselines with MixTTA to the low-rank parameterization rank $r$ on ImageNet-C at severity level~5. 
We observe that performance improves as $r$ increases from 1 to 4, reaching the best average accuracy at $r=4$, indicating that a modest rank is sufficient to capture the dominant cross-channel shift required at test time. 
However, further increasing the rank to 8 or 16 degrades accuracy, suggesting that overly expressive channel mixing introduces unnecessary degrees of freedom and leads to less stable adaptation. 
At the limit, the results demonstrate that replacing the diagonal affine scale matrix $\Gamma$ in Eq.~\eqref{eqn:tent_matrix} with a full channel-mixing transform $W\in\mathbb{R}^{C\times C}$ (``Full rank'') causes catastrophic collapse. We attribute this to the substantially larger parameterization, which weakens the inductive bias of affine-style modulation and leads to unstable online optimization and severe overfitting. 
These results collectively suggest that the low-rank parameterization effectively regularizes the update dynamics, maintaining robustness under severe corruption.
Notably, $r=4$ yields the best accuracy across all five baselines simultaneously, eliminating the need for per-method hyperparameter tuning.

\begin{table}[t]
\centering
\small
\setlength{\tabcolsep}{2mm}
\resizebox{0.9\linewidth}{!}{%
\begin{tabular}{l|c|c|c}
\textbf{Method} & \textbf{\# Params} & \textbf{GPU Time (50,000)} & \textbf{Acc. (\%)} \\
\hline
DeYO     & 27,648 & 477 seconds & 63.8 \\ 
DeYO (Full block)   & 36,864 & 481 seconds & 62.1 \\
\rowcolor{lightgray}
DeYO+MixTTA    & 46,080 & 496 seconds & \textbf{65.6} \\ 
\hline
ReCAP    & 27,648 & 451 seconds & 64.0 \\ 
ReCAP (Full block)  & 36,864 & 455 seconds & 62.4 \\ 
\rowcolor{lightgray}
ReCAP+MixTTA   & 46,080 & 470 seconds & \textbf{64.9} \\ 
\end{tabular}%
}
\caption{
Comparison of learnable parameter counts, runtime, and accuracy for DeYO, DeYO+MixTTA, ReCAP, and ReCAP+MixTTA, respectively. 
We also report baselines extended to update all transformer encoder blocks.
The practical runtime is evaluated using a single A5000 GPU.
}
\label{tab:param_comparison}
\end{table}

\subsubsection{Overhead and Parameter-controlled Analysis.}
Table~\ref{tab:param_comparison} reports the learnable parameter count, GPU running time (50,000 samples), and average accuracy on the ImageNet-C dataset at severity level 5 of DeYO and ReCAP with and without MixTTA. 
MixTTA adds approximately 4\% runtime overhead for both DeYO and ReCAP, confirming that the additional computation is negligible in practice.

We further disentangle the effect of MixTTA from simply increasing the number of learnable parameters by comparing against the full-block update variants of DeYO and ReCAP. 
Naively extending adaptation to all transformer encoder blocks increases the parameter count but degrades performance for both methods, suggesting that deeper online updates can be unstable or prone to overfitting. 
In contrast, integrating MixTTA consistently improves these baselines, indicating that the gains stem from the structured low-rank cross-channel transformation of MixTTA rather than from simply increasing model capacity.

\begin{figure*}[t]
  \centering
  \begin{subfigure}[b]{0.32\linewidth}
      \includegraphics[width=\linewidth]{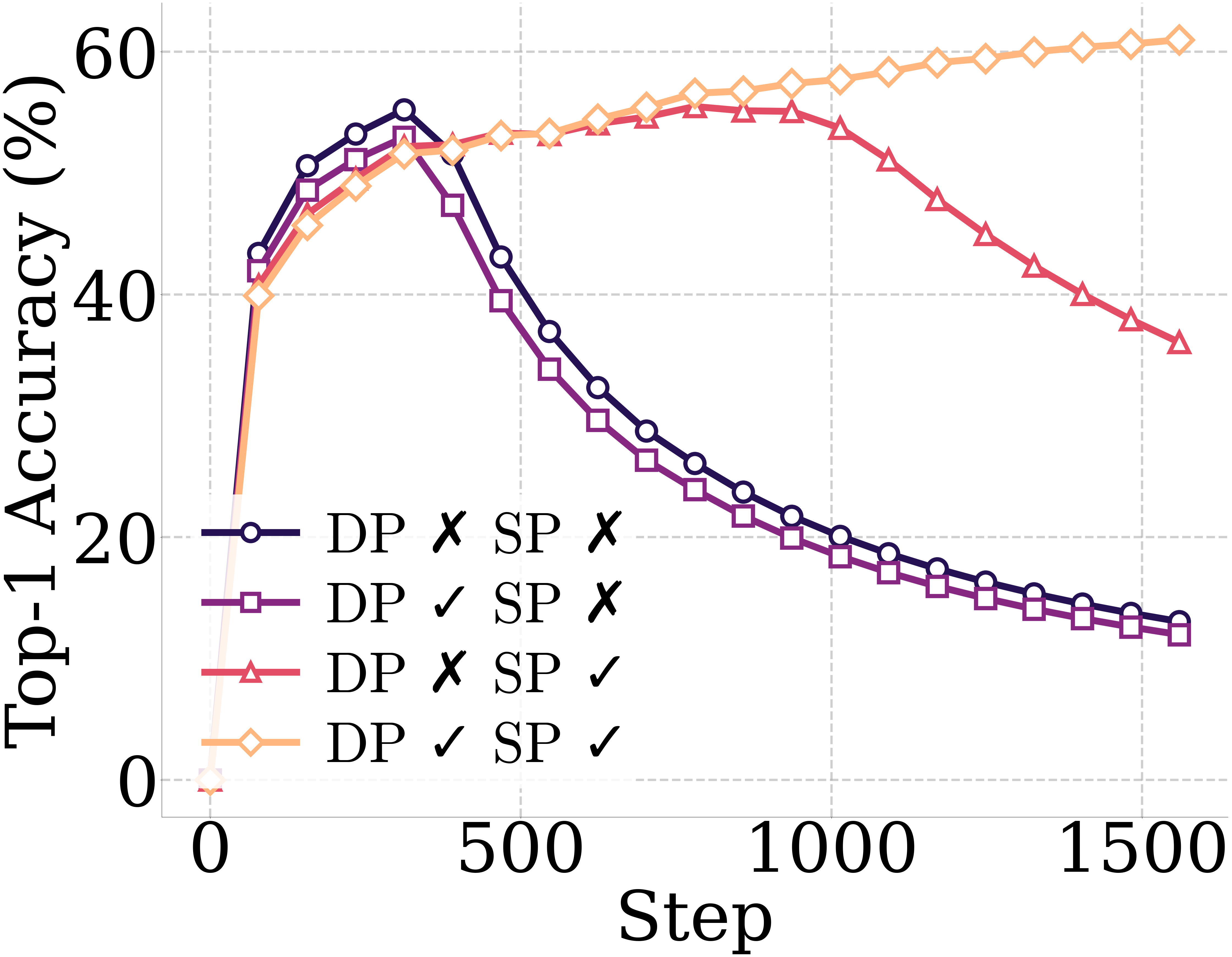}
      \caption{Accuracy}
      \label{fig:acc_plot}
  \end{subfigure}
  \begin{subfigure}[b]{0.32\linewidth}
      \includegraphics[width=\linewidth]{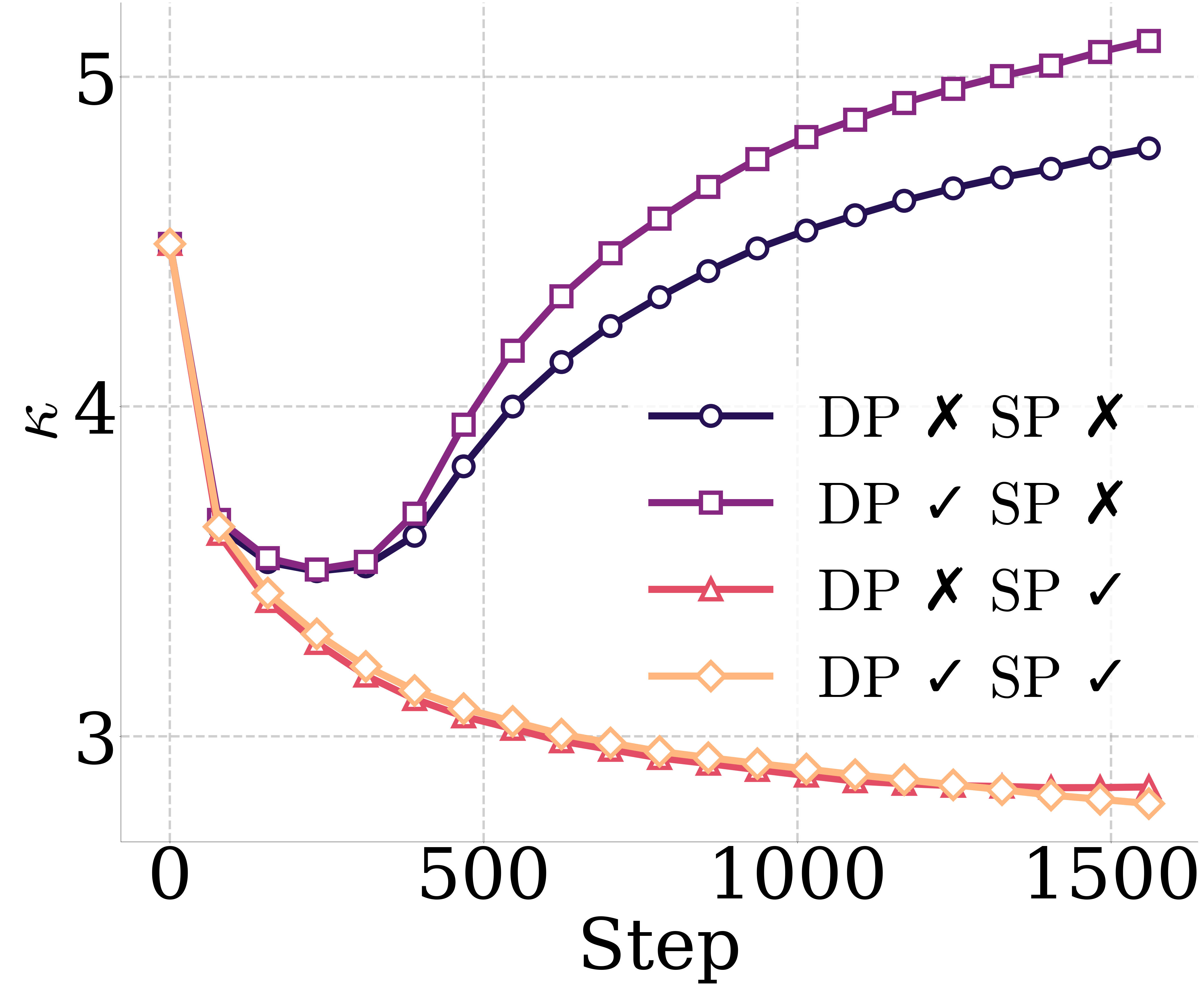}
      \caption{$\kappa$}
      \label{fig:kappa_plot}
  \end{subfigure}
  \begin{subfigure}[b]{0.32\linewidth}
      \includegraphics[width=\linewidth]{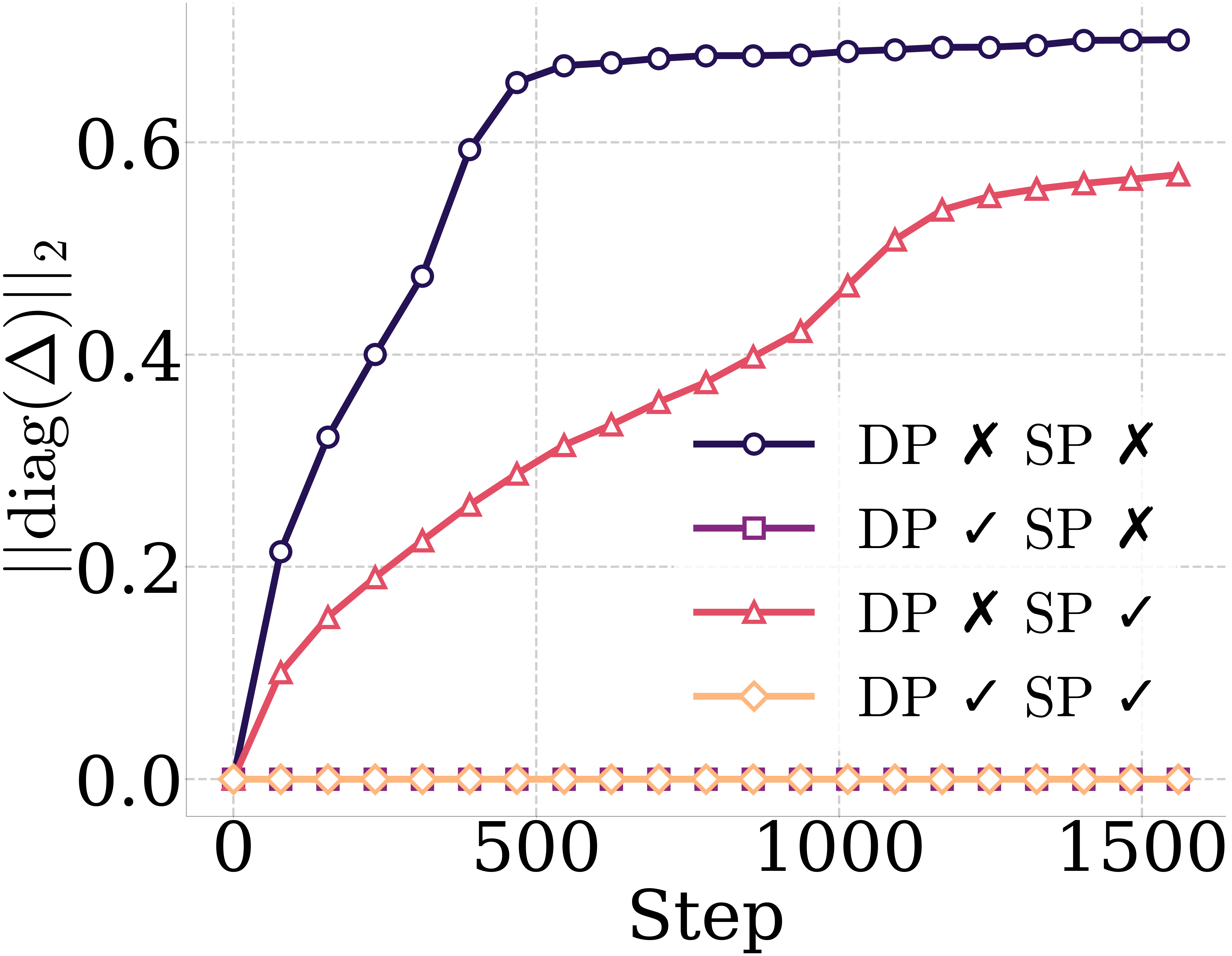}
      \caption{$||\mathrm{diag}(\Delta)||_2$}
      \label{fig:diag_plot}
  \end{subfigure}
    \caption{
    Comparison of (a) top-1 accuracy, (b) the condition number $\kappa$ of the feature covariance at an intermediate layer, and (c) $||\operatorname{diag}(\Delta)||_2$ for adapted models on ImageNet-C \emph{snow} (severity 5) under imbalanced label shift, across different projection configurations. 
    }
\label{fig:ablation}
\end{figure*}

\subsubsection{Ablation Study of Projection Strategies.}
To examine the individual efficacy of Decoupling Projection (DP) and Spectral Projection (SP), we track accuracy, feature covariance condition number $\kappa=\lambda_{\max}/\lambda_{\min}$, and the norm of the diagonal component $\|\operatorname{diag}(\Delta)\|_2$ over adaptation steps on ImageNet-C \emph{snow} corruption at severity level~5 under imbalanced label shift, across all DP/SP configurations.

Fig.~\ref{fig:acc_plot} shows that enabling both DP and SP yields the most stable accuracy trajectory, preventing the performance collapse observed in other configurations. 
Particularly, when SP is disabled, accuracy begins to drop sharply, and this transition aligns with the rapid growth of the condition number in Fig.~\ref{fig:kappa_plot}, indicating that the failure coincides with a rank-1-dominant covariance regime where the leading eigenvalue overwhelms the spectrum. 
In contrast, enabling SP keeps $\kappa$ bounded and thereby stabilizes the adaptation dynamics, explaining its critical role in preventing collapse.

Consistent with the design goal of DP, enabling DP effectively suppresses diagonal leakage in the low-rank branch, yielding a markedly smaller $\|\operatorname{diag}(\Delta)\|_2$ than DP-off configurations (Fig.~\ref{fig:diag_plot}). 
When DP is disabled, $\|\operatorname{diag}(\Delta)\|_2$ grows progressively over adaptation steps, indicating increasing diagonal leakage. 
In the DP-off/SP-on setting, accuracy degradation coincides with a drastic rise in $\|\operatorname{diag}(\Delta)\|_2$. This suggests that diagonal leakage may partially undermine the intended cross-channel mixing effect of the low-rank branch by reintroducing per-channel modulation. 


\subsubsection{Analysis of Correlation Shifts.}
\begin{figure*}[t]
  \centering
  \begin{subfigure}[b]{0.48\linewidth}
      \includegraphics[width=\linewidth]{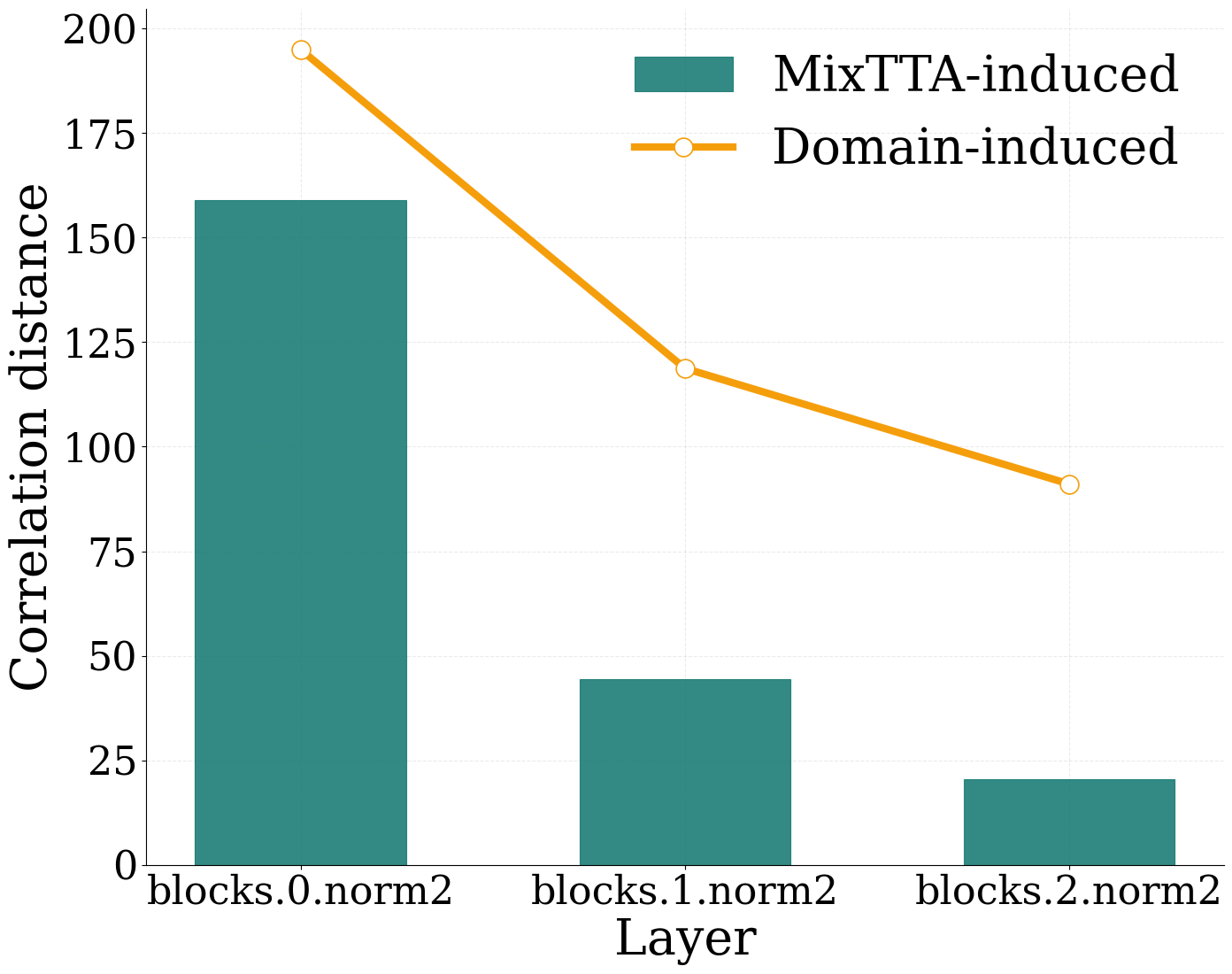}
      \caption{Correlation Distance}
      \label{fig:layer_wise_corr}
  \end{subfigure}
  \begin{subfigure}[b]{0.48\linewidth}
      \includegraphics[width=\linewidth]{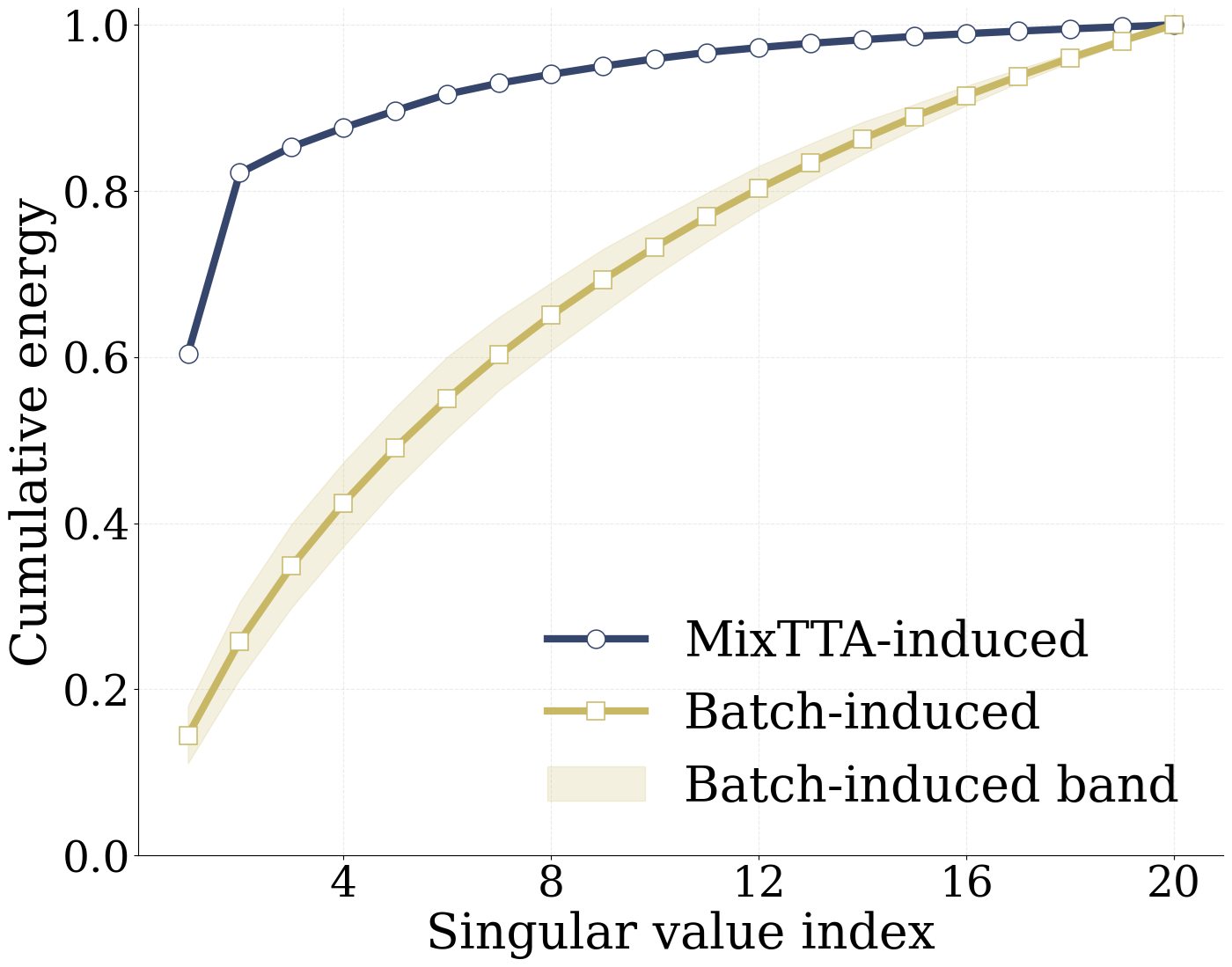}
      \caption{Cumulative Energy}
      \label{fig:cumulative_energy}
  \end{subfigure}
  \caption{
    Layer-wise and spectral analysis of feature correlation changes on ImageNet-C \emph{Gaussian noise} at severity level~5.
    (a) Layer-wise correlation distances. 
    (b) Cumulative energy of the singular-value spectrum at \texttt{blocks.0.norm2}.
    }
\label{fig:corr_shift_analysis}
\end{figure*}

To examine how MixTTA modifies cross-channel correlations during adaptation, we compute the correlation changes on ImageNet-C \emph{Gaussian noise} at severity level~5.
In Fig.~\ref{fig:layer_wise_corr}, we visualize two layer-wise quantities: the MixTTA-induced shift, computed as the correlation distance between Tent and Tent+MixTTA on the same corrupted input, and the domain-induced shift, computed as the correlation distance between source and corrupted features under the source model. 
Although these two correlation shifts have different origins, both quantities peak at the earliest block and decrease in deeper blocks, indicating that MixTTA applies larger corrections at layers where correlation disruption is most severe.
Notably, this layer-wise allocation arises purely from entropy-driven optimization, without any explicit layer-dependent weighting, suggesting that the low-rank branch adaptively responds to cross-channel misalignment across layers.

Fig.~\ref{fig:cumulative_energy} further examines the spectral structure of the MixTTA-induced correlation shift. 
Specifically, we plot how rapidly the cumulative spectral energy of the MixTTA-induced correction concentrates as a function of singular-value index.
As a control, we estimate batch-induced variation by randomly splitting a target mini-batch into two halves and measuring the correlation difference between the two half-batch means; we report its mean curve together with the $q{=}0.1$--$0.9$ quantile band across random splits.
Compared to this batch-noise baseline, the MixTTA-induced correction concentrates energy in markedly fewer singular directions, forming a structured, low-rank adjustment rather than a diffuse perturbation. This is consistent with the intended design of the low-rank module.

\section{Conclusion}

We identified that distribution shift disrupts inter-channel feature correlations in a way that conventional per-channel affine updates cannot correct, and that this disruption is most severe in earlier network layers. Based on this finding, we proposed MixTTA, a plug-in module that extends normalization layers with a low-rank cross-channel transformation, enabling inter-channel mixing across multiple layers. 
Together with Decoupling Projection to enforce strict diagonal/off-diagonal separation and Spectral Projection to prevent rank-1 collapse,
MixTTA consistently improves existing normalization-based TTA methods across standard and wild scenarios on ImageNet-C and ImageNet-Sketch, while incurring only minimal overhead.
One future direction is to develop an adaptive rank selection strategy that adjusts the subspace dimension to the severity and structure of the encountered domain shift.
We hope that the perspective of cross-channel correlation modeling opens new directions for more expressive and reliable test-time adaptation in the community.

\subsubsection{Acknowledgements.} 
This work was supported by the Institute of Information \&Communications Technology Planning \& Evaluation (IITP) under the AI Star Fellowship (Kookmin University) (RS-2025-02219317 (10\%)) and the Leading Generative AI Human Resources Development (IITP-2026-RS-2026-25546026 (10\%)) grant funded by the Korea government (MSIT), and the ``Advanced GPU Utilization Support Program'' funded by the Government of the Republic of Korea (Ministry of Science and ICT), the National Research Foundation of Korea (NRF, RS-2024-00451435 (10\%), RS-2024-00413957 (10\%)), the Institute of Information \& Communications Technology Planning \& Evaluation (IITP, RS-2025-02305453 (15\%), RS-2025-02273157 (15\%), RS-2025-25442149 (15\%), RS-2021-II211343 (15\%)) grant funded by the Ministry of Science and ICT (MSIT), the Institute of New Media and Communications (INMAC), and the BK21 FOUR program of the Education, the Artificial Intelligence Graduate School Program (Seoul National University), and the Research Program for Future ICT Pioneers, Seoul National University in 2026.

%
%

\bibliographystyle{splncs04}
\bibliography{main}
\end{document}